\documentclass[twocolumn]{article}
\usepackage{graphicx} 
\usepackage{hyperref}
\usepackage{xurl}
\usepackage{cleveref}
\usepackage[a4paper,
            bindingoffset=0.2in,
            left=1in,
            right=1in,
            top=1.4in,
            bottom=1in,
            footskip=.25in]{geometry}
\usepackage{custom_commands}

\date{} 

\title{Improving Detection of Person Class Using Dense Pooling}
\author{\capitalisewords{N}ouman \capitalisewords{A}hmad  $^{1}$
,
}

\begin{document}
\twocolumn[
  \begin{@twocolumnfalse}
    \maketitle
        \begin{abstract}
Lately, the continuous development of deep learning models by many researchers in the area of computer vision has attracted more researchers to further improve the accuracy of these models. FasterRCNN \cite{ren2015faster} has already provided a state-of-the-art approach to improve the accuracy and detection of 80 different objects given in the COCO dataset. To further improve the performance of person detection we have conducted a different approach which gives the state-of-the-art conclusion. An ROI is a step in FasterRCNN that extract the features from the given image with a fixed size and transfer into for further classification. To enhance the ROI performance, we have conducted an approach that implements dense pooling and converts the image into a 3D model to further transform into UV(ultra Violet) images which makes it easy to extract the right features from the images. To implement our approach we have approached the state-of-the-art COCO datasets and extracted 6982 images that include a person object and our final achievements conclude that using our approach has made significant results in detecting the person object in the given image.  
\end{abstract}
\hspace{5pt}


\hspace{5pt}
        \end{@twocolumnfalse}
]
\section{Introduction}
Detecting the object from an image is a crucial problem. Many researchers have conducted different approaches to address this issue and presented their solutions by adapting the unique experiments \cite{huang2019few}\cite{lee2018stacked}\cite{li2019visual}. The precision of such experiments is being enhanced. Normally, an image is given to the system, and the system analyzes the image and provides the final image by labeling the detected objects. This process seems smooth but to accomplish this objective is not that linear. The success of these tasks relies on how well your model was trained and how quickly it can detect the input and return the result. Since detecting an object from an image depends on its training when an image is given as input that is why choosing the right model is important.

Recently, many models have been proposed. However, Faster-RCNN \cite{ren2015faster} is an individual-stage model that is up-skilled end-to-end. It manipulates a novel region proposal network (RPN) for causing region proposals, which optimizes one's time and increases efficiency while weights up to traditional algorithms. It manipulates the ROI Pooling layer to draw out a fixed-length feature vector from each region proposal. The RPN is lined up end-to-end to produce high-quality region proposals, which are pre-owned by Fast R-CNN\cite{girshick2015fast} for detection. RPN and Fast R-CNN are united into a single network by sharing their convolutional networks and providing the final state-of-the-art result.

Mask RCNN \cite{he2017mask} is an alternative choice to train the model on the COCO dataset \cite{lin2014microsoft}. However, it has the disadvantage of identifying the object suffered from motion blur at low resolution as hand \cite{8593130}. Since dense pooling involves transferring images into 3D dimension \cite{neverova2020continuous} that is why we have chosen Faster RCNN \cite{ren2015faster} for our experiments.

To further investigate object detection, the dense pooling \cite{wu2019detectron2}technique was proposed and it plunges into the tasks of determining and describing dense concurrence in deformable object categories. However, these earlier attempted complications in the form of suggested solutions are still at an ad-hoc level for specific object types (for example, humans). Nevertheless, scaling the geometry manipulation to all human objects in real scenarios requires those techniques that can tell us the inner correspondences between all possible layers inside the different objects and dense pooling is one of them that transforms each pixel in a 2D image which is an embedding vector representation of the related vertex in the object mesh. However, to accomplish the dense representation between image pixels and 3D object geometry is the main target of this approach which causes the filtration of under detection object and is used to extract the object.

With continuous improvement in model development one straightforward way comes in mind to introduce a new backbone then replace the large backbone networks with the new ones and use the N-to-N experiments to achieve a more accurate model. However, after extensive experiments we find that directly training a small model from scratch cannot achieve high accuracy, due to its limited model functionalities. Knowledge Distillation \cite{hinton2015distilling} is a very beneficial way, which can advance the precision of the new model by leveraging knowledge from the main
model. Apart from available existed approaches \cite{chen2017learning}, \cite{chen2020imram} that are proposed for filtering a single 1D vector explaining the object detection in the given image, we put our attention on purifying the image by transferring into vector shape so that a complete outcomes can be achieved without any loss from the source.

Faster-RCNN without dense pooling is able to detect human class. However, merging both approaches boosts accuracy. In our study, we have utilized the Faster-RCNN Resnet-50 RPN and Resnet-101 with dense pooling and experimented using COCO dataset\cite{lin2014microsoft}. Since our study currently focuses only on human detection and not all the classes that Faster-RNCC is already able to perform that is why we are currently taking the 6982 images from the COCO dataset and training our model on a GPU of type T4 with 16GB RAM. Our Results show that the proposed model is able to detect human objects more precisely than detecting via Faster RCNN without using dense pooling. 

Our attempted experiments with special attention as explained in \cref{fig_01} on improving the accuracy of the Faster RCNN model have been made available open source on a given resource. In the open source, we have used a recommended resource to obtain the COCO dataset by their official source i.e. Fiftyone\cite{moore2020fiftyone}. This resource is feasible in taking into consideration multiple use cases while using the COCO dataset according to ones own unique need. Further we have compared our results with Yolo-v7 \cite{wang2023yolov7} and found that our experiments provide state-of-the-art results.

\url{
https://github.com/NoumanAhmad448/Improving\_Detection\_of\_Person\_using\_dense\_pooling}
\section{Related Work}
\subsection{Object Detection}
Several versatile adaptable algorithms and techniques have been advanced for images and the correctness has been significantly enhanced. There is a depth of work on object detection methods. An inclusive survey and proposed models on object detections can be discovered in
\cite{hosang2014good}, \cite{hosang2015makes}, \cite{chavali2016object}. These approaches are either based on grouping super-pixels or sliding windows.

Those approaches that are based on super-pixels are
Selective Search \cite{uijlings2013selective}, CPMC \cite{carreira2011cpmc}, MCG \cite{arbelaez2014multiscale}) and those
that are based on sliding windows are objectness in windows
\cite{alexe2012measuring}, EdgeBoxes \cite{zitnick2014edge}. These comprehensive and inclusive Object identification approaches were chosen as external modules completely independent of the detectors i.e.  RCNN [5], object detectors Selective Search \cite{uijlings2013selective} Fast R-CNN \cite{girshick2015fast} and to improve their accuracy Faster RCNN \cite{ren2015faster} method and Mask RCNN \cite{he2017mask} are proposed.

\subsection{Proposed Models}
 Previously proposed R-CNN approach\cite{girshick2014rich} 
 takes the input and trains it end-to-end using CNNs layers to classify the proposal regions into object recognition and classify it. It proposes the technique as a classifier however, it does not anticipate object bounds but it does not true for purifying by bounding box regression. Its comprehensive precision is analyzed on the performance of the region proposal module. It can further be checked on \cite{hosang2015makes}. Many proposed distinct approaches have given the idea of using deep learning networks for predicting object bounding boxes \cite{szegedy2013deep}, \cite{sermanet2013overfeat}, \cite{erhan2014scalable}, \cite{szegedy2014scalable},\cite{wang2023yolov7}. 
 
According to the survey conducted by \cite{yuan2023comprehensive},\cite{zou2023object},\cite{oksuz2020imbalance},\cite{zaidi2022survey},\cite{wang2023multi} many state-of-the-art approaches have been advanced and their accuracy were continuously enhanced. Many researchers have found the Yolo \cite{wang2023yolov7} as the state-of-the-art approach and they have proposed many approaches by continuously increasing the accuracy in the depth of object detection \cite{mahaur2023small}. Most of these approaches are addressed to solve the autonomous driving problems \cite{wang2023sat}. Along with suggestions of many state-of-the-art approaches, some researchers have put their effort to enhance the dataset to provide more opportunities to perform research more linearly \cite{meng2023detection}. Some researchers have proposed algorithms to detect objects for multiple datasets \cite{zhao2022omdet}.
 
 In the OverFeat approach \cite{sermanet2013overfeat}, a fully-connected layer is modified to propose the box synchronization for the localization object that presumes a single object to be detected. It is then promoted to the next step to recognize multiple classification classes. The MultiBox methods \cite{erhan2014scalable}, \cite{szegedy2014scalable} are advanced for the prediction of multiple class-agnostic boxes and conclude the “singlebox” fashion of OverFeat. These approaches
are used as proposals for R-CNN \cite{girshick2014rich}. The disadvantage of using MultilBox is that it does not provide corporations with sharing features between the proposal and detection networks. Further it only is applicable on a single cropped image or multiple large cropped images, in general, 224×224 size. Both MultiBox and OverFeat are later enhanced with the approach of Faster RCNN \cite{ren2015faster}. Further, Continuous growth of DeepMask approach \cite{o2015learning} and other advance approaches \cite{sermanet2013overfeat}, \cite{he2015spatial}, \cite{dai2015convolutional},
\cite{long2015fully}, \cite{girshick2015fast} have attracted the modified and increasing efficient. 
For efficient and enhanced region-based semantic segmentation and object detection adaptively-sized pooling (SPP) \cite{he2015spatial} was brought to the future research and shares the
convolutional feature maps which are developed \cite{dai2015convolutional}. Fast R-CNN \cite{girshick2015fast} brings 
\begin{table*}[h]
\centering
\def\arraystretch{1.5}%
\setlength{\tabcolsep}{28pt}
\setlength{\arrayrulewidth}{0.5mm}
\caption{This implementation is performed on COCO dataset 
 with 100 epochs without the usage of dense pooling.}
\begin{tabular}{|p{0.8cm}|p{0.8cm}|p{0.8cm}|p{0.8cm}|p{0.8cm}|p{0.5cm}| }
 \hline
 \multicolumn{6}{|c|}{\textbf{Implementation of Faster RCNN without Dense Pooling}} \\
 \hline
  \textbf{AP}   &  \textbf{AP50}  &  \textbf{AP75}  &  \textbf{APs}  &  \textbf{APm}  &  \textbf{APl}  \\
 \hline
   47.9 & 80.9  & 52.6  & 23.9 & 50.3 & 67.5 \\
 \hline
\end{tabular}
 
\label{table_01}
\end{table*}

\begin{table*}[h]
\centering
\def\arraystretch{1.5}%
\setlength{\tabcolsep}{28pt}
\setlength{\arrayrulewidth}{0.5mm}
 \caption{This implementation is performed on COCO dataset 
 with 100 epochs with the usage of dense pooling and Faster RCNN ResNeXt50.}
\begin{tabular}{|p{0.8cm}|p{0.8cm}|p{0.8cm}|p{0.8cm}|p{0.8cm}|p{0.5cm}| }
 \hline
 \multicolumn{6}{|c|}{\textbf{Implementation of Faster RCNN ResNeXt50 with Dense Pooling}} \\
 \hline
 \textbf{AP}   &  \textbf{AP50}  &  \textbf{AP75}  &  \textbf{APs}  &  \textbf{APm}  &  \textbf{APl}  \\
 \hline
   49.9  & 86.1  & 52.8  & 26.2 & 51.5 & 68.9 \\
 \hline
\end{tabular}
\label{table_02}
\end{table*}
detector training on shared convolutional features and complies with accuracy and speed. Faster RCNN \cite{ren2015faster}brings the new concept of RPN(region proposal network) and provides the solution of object detection using RPN to generate the region proposals. Further, Mast Faster RCNN \cite{he2017mask} adeptly recognizes objects in an image while at the same time generating a high-quality segmentation mask for each instance. Continuous Surface Embeddings \cite{neverova2020continuous} is an approach for transforming each pixel in a 2D image into 3D by embedding a vector of the corresponding vertex in the object mesh while establishing dense correspondences. Finally, recent work  \cite{neverova2021discovering} has shown that dense posing is possible for several objects.
\section{Methodology}
Dense posing is a process of establishing a dense correspondence between a surface-based representation and an RGB image of the human body and a dense correspondence is a mapper of transforming the parts of images into another image for detection or classification purpose \cite{guler2018densepose}. However, accomplishing this correspondence requires manual work \cite{guler2018densepose}. To overcome the manual approach, we use the automatic method \cite{neverova2020continuous} which is defined in \cref{main_eq}

\begin{equation}
\label{main_eq}
p(X|x, y, e, \psi ) =\frac{
\exp(-e_{X}, \psi_{x}(y))}{
                    \displaystyle \int  _{\beta}(\exp(-e_{X}, \psi_{x}(y)) dX)}
\end{equation}

\begin{figure}[hbt!]
\centering
\includegraphics[width=0.50\textwidth]{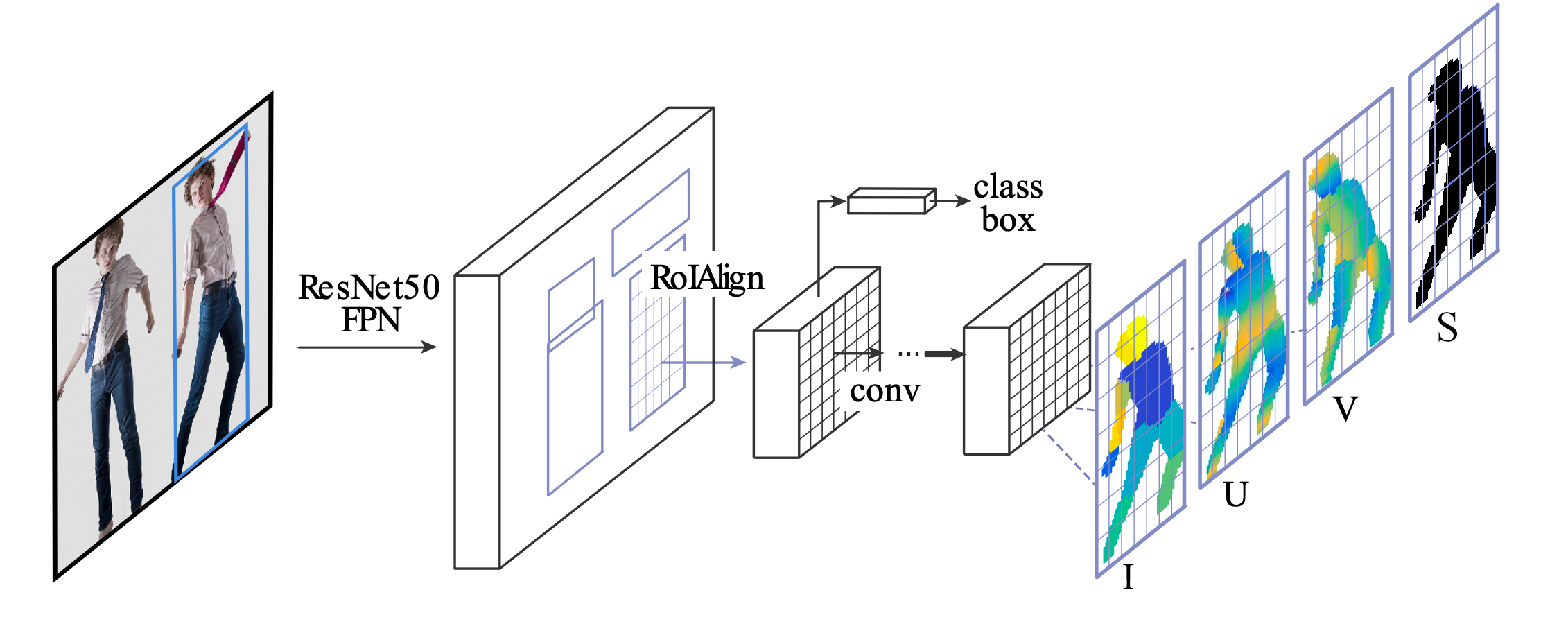}
\caption{DensePose design based on Faster R-CNN with F-50 Feature Pyramid Network (FPN).}
\label{fig_01}
\end{figure}

In \cref{main_eq}, the embedding function $e_{X}$ represents a learnable parameter similar to the network. While $\beta$ represents the surface with a mesh, $\psi$ is a deep network that needs to be trained with the surface. The basic approach of experimenting the dense pooling is to find the approximate surface $\beta$ with a mesh and vertices to point out the main regions of an image that are desired to generate the image that is used for detection.

\begin{equation}
\label{eq2}
    L(E, \psi) =  \log{(k|x,\beta,E,\psi)}
\end{equation}
Our suggested \cref{main_eq} model is a simplified explanation because it needs the initial mesh to further process the image while the embedding matrix E is taken in automatically. This minimizes the manual workload that is needed to instantiate densePose \cite{neverova2020continuous}. It reduces the work and makes the model efficient to filter the image for detection purposes under the given circumstances. The overall approach is summarized in \cref{fig_01} that include selecting the image as input and transforming into another image that is later used to detect the object.
\begin{table*}[h]
\centering
\def\arraystretch{1.5}%
\setlength{\tabcolsep}{28pt}
\setlength{\arrayrulewidth}{0.5mm}
 \caption{This implementation is performed on COCO dataset 
 with 100 epochs without the usage of dense pooling while using the Resnet-101 backbone.}
\begin{tabular}{|p{0.8cm}|p{0.8cm}|p{0.8cm}|p{0.8cm}|p{0.8cm}|p{0.5cm}| }
 \hline
 \multicolumn{6}{|c|}{\textbf{Implementation of Faster RCNN without Dense Pooling with backbone Resnet-101}} \\
 \hline
 \textbf{AP}   &  \textbf{AP50}  &  \textbf{AP75}  &  \textbf{APs}  &  \textbf{APm}  &  \textbf{APl}  \\
 \hline
  59.0 & 93.4  & 67.9  & 35.3 & 61.9 & 78.2 \\
 \hline
\end{tabular}
\label{table_03}
\end{table*}

\begin{table*}[h]
\centering
\def\arraystretch{1.5}%
\setlength{\tabcolsep}{28pt}
\setlength{\arrayrulewidth}{0.5mm}
 \caption{This implementation is performed on COCO dataset 
 with 100 epochs while using dense pooling and training on Resnet-101 backbone.}
\begin{tabular}{|p{0.8cm}|p{0.8cm}|p{0.8cm}|p{0.8cm}|p{0.8cm}|p{0.5cm}| }
 \hline
 \multicolumn{6}{|c|}{\textbf{Implementation of Faster RCNN while using Dense Pooling and backbone Resnet-101}} \\
 \hline
   \textbf{AP}   &  \textbf{AP50}  &  \textbf{AP75}  &  \textbf{APs}  &  \textbf{APm}  &  \textbf{APl}  \\
 \hline
  55.2 & 95.1  & 59.0  & 30.6 & 58.9 & 73.4 \\
 \hline
\end{tabular}
\label{table_04}
\end{table*}
\section{Experiments And Results}
To experiment with the Faster RCNN with the dense pooling approaches, we have reached the GPU system T4 with 15GB RAM and trained the model under 100 epochs on the COCO dataset. We have trained two different models on the COCO dataset. One model is 

trained without the usage of dense pooling to recognize the human class while the other is trained with the dense pooling as shown in \cref{table_01}. We have used two different backbones for training the dataset. One model is trained on resnet-50 shown in \cref{table_01},\cref{table_02} while the other is trained on resnet-101 backbone shown in \cref{table_03}, \cref{table_04}. Our conducted experiments using backbone resnet-50 have found better results than the model that is trained without dense pooling and with the same backbone. However, in training the model with backbone resnet-101 we have observed that at level 50, resnet-101 gives a better result than the model that was trained without dense pooling while using the same backbone. Furthermore, it can be seen in \cref{table_04} the results are not bad and their accuracy can further be increased while adopting different settings and training on larger dataset. It is better to mention that we have trained the model with backbone-101 on 300 epochs and found different results. It is recommended to train our model on 100 epochs. 

Although both approaches (i.e. Faster RCNN and Dense Pooling) are for object detection and are being used widely on large scales however our experiments reveal that merging both techniques provide state-of-the-art results as shown in \cref{table_02}. 
\begin{figure}[hbt!]
\centering
\includegraphics[width=0.50\textwidth]{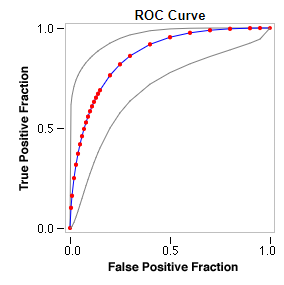}
\caption{Area under ROC curve w.r.t. false positive and true positive values chosen after training the model with backbone of resnet-50 and resnet-101 }
\label{fig_03}
\end{figure}
MS COCO \cite{lin2014microsoft} contains 118K images for training while 5K for validation. but these images include multiple classes e.g. dogs, cats, etc. since our experiments are currently focusing on a human class which is why we have extracted 6982 images and trained our models on them.Further, our experiments focus detecting the images that include human class.

\begin{table}[h!]
\centering
\def\arraystretch{1.5}%
\setlength{\tabcolsep}{15pt}
\setlength{\arrayrulewidth}{0.3mm}
 \caption{Comparison between our proposed method and state-of-the-art methods.}
\begin{tabular}{|p{1.5cm}|p{0.5cm}|p{0.5cm}|p{0.5cm}| }
 \hline
 \multicolumn{4}{|c|}{\textbf{Precision Scales} } \\
 \hline 
 \textbf{Method} & \textbf{AP} & \textbf{AP50} & \textbf{AP75} \\ 
 \hline
 AMA-net \cite{guo2019adaptive} & 64.1 & 91.4 & 72.9\\
 \hline
DensePose \cite{guler2018densepose} & 66.4 & 92.9 & 77.9\\
\hline
DensePose-DeepLab \cite{guler2018densepose} & 51.8 & 83.7 & 56.3  \\
\hline
CSE \cite{neverova2020continuous} & 67.0 &  93.8 & 78.6 \\
\hline
CSE-DeepLab \cite{neverova2020continuous} &  68.0 & 94.1 &  80.0 \\
\hline
BodyMap RGB-only \cite{ianina2022bodymap} & 71.0 & 94.3 & 83.3 \\
  \hline
Yolo-v7 \cite{wang2023yolov7} & 51.2 & 69.7 & 55.5\\
  \hline
  Our Method & 55.2 & 95.1  & 59.0  \\
 \hline
\end{tabular}
\label{table_05}
\end{table}
Recently proposed approaches \cite{neverova2021discovering}, \cite{neverova2020continuous} uses the dense pooling to detect the object class. However, it is found that \cite{neverova2021discovering} only focuses on identification animal class and their accuracy at AP, AP-50, AP-75 are found 37.5,67.8,36.4 respectively. Similarly, we have compared our results with \cite{neverova2020continuous} and found that our proposed model provide the relatively better accuracy. As \cite{neverova2020continuous} describes training the inner model with dense pooling approach requires manual work which cause quality deficiency and require time too to perform this work. We have approached the same idea and trained our model without any manual requirement.

BodyMap \cite{ianina2022bodymap} is a recently proposed method on top of dense pooling. However, when it comes to choosing the RGB images it lacks in performance and our proposed results are found more accurate. \cite{ianina2022bodymap},\cite{neverova2020continuous},\cite{guo2019adaptive} are those methods that are earlier proposed however, they lack in performance of identifying the human pose. \cref{table_05} 

\begin{figure}[hbt!]
\centering
\includegraphics[width=0.50\textwidth]{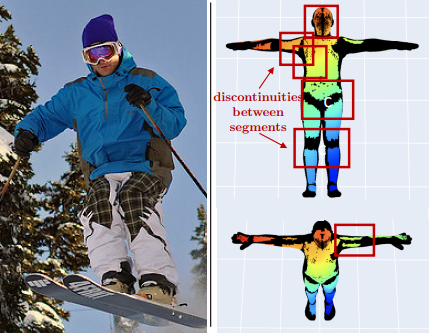}
\caption{Identification of human object using densePose by detection the body parts as segment}
\label{fig_02}
\end{figure}

\begin{table}[h!]
\centering
\def\arraystretch{1.2}%
\setlength{\tabcolsep}{16pt}
\setlength{\arrayrulewidth}{0.3mm}
 \caption{Summary on AP(Average Precision) AND AR(average Recall) based on backbone chosen with and without dense pooling and training on COCO dataset \cite{lin2014microsoft}. 
*BB represents backbone. These values are calculated on IoU=0.50:0.95, area=large and epoch=100
 }
\begin{tabular}{|p{1.5cm}|p{0.5cm}|p{0.5cm}|p{0.5cm}| }
 \hline
 \multicolumn{4}{|c|}{\textbf{AP AND AR VALUES }} \\
 \hline 
 \textbf{Model} & \textbf{BB*} & \textbf{Type} & \textbf{Value} \\
 \hline
 Without Dense Pooling & 50 & AP & 7.0 \\
 \hline
 Without Dense Pooling & 50 & AR & 8.0 \\
 \hline
 With Dense Pooling & 50 & AP & 7.0 \\
 \hline
 With Dense Pooling & 50 & AR & 8.0 \\
 \hline
 Without Dense Pooling & 101 & AP & 8.0 \\
 \hline
 Without Dense Pooling & 101 & AR & 9.0 \\
 \hline
 With Dense Pooling & 101 & AP & 7.0 \\
 \hline
 With Dense Pooling & 101 & AR & 8.0 \\
  \hline
\end{tabular}
\label{table_06}
\end{table}

explains the comparison of earlier proposed approaches and our proposed method. Our proposed model uses dense pooling as shown in \cref{fig_02}. Our Further, we have calculated the AP(average precision) and AR(average recall) at IOU=0.50:0.90 while training the model with and without dense pooling \cref{table_06}. These values represent that proposed model works well in the object detection. Further, we have calculated AUC curve for False Positive and True Positive values \cref{fig_03} which shows the accuracy of 84\% with resnet-50 and resnet-101.

\section{Conclusion} Detecting the object is a common problem. Recently it is observed that many approaches have been proposed and their accuracy has significantly improved. Faster RCNN \cite{ren2015faster} is the state-of-the-art approach that is commonly used to detect multiple classes and it is trained on both coco 2014 and 2017 datasets \cite{lin2014microsoft}. \cite{guler2018densepose} is an approach that is used to identify the human body parts by transforming the image into another image which is used to detect the object. In this paper, we have approached the combination of both approaches and presented our experiments with the backbone of Resnet50 and trained the model on GPU T4 with the coco 2017 dataset \cite{lin2014microsoft} with 100 epoch and after the extensive experiments we have found that our results reveal that utilizing the combination of both approaches together has better consequences than training the model only on Faster RCNN \cite{ren2015faster} without the dense pooling. Our experiments are currently targeted to improve the human class only. However, in future experiments, we will take our analysis to the next level by training the model on animal classes while using both Resnet50 and Resnet101 backbone with dense pooling and Faster RCNN or Mask Faster RCNN \cite{he2017mask} and improving their accuracy with the state-of-the-art results.
\bibliographystyle{plain}
\bibliography{bibo}
\end{document}